\documentclass[runningheads]{llncs}

\usepackage[T1]{fontenc}
\usepackage{framed,multirow}
\usepackage{amssymb}
\usepackage{latexsym}
\usepackage{url}
\usepackage{xcolor}
\usepackage{times}
\usepackage{epsfig}
\usepackage{graphicx}
\usepackage{amsmath}
\usepackage{amssymb}
\usepackage{subcaption}
\usepackage{booktabs}
\usepackage{threeparttable}
\usepackage{algorithmic}
\usepackage[linesnumbered,ruled,vlined]{algorithm2e}
\usepackage{hyperref}
\usepackage{colortbl}
\usepackage[numbers]{natbib}
\usepackage{setspace}

\definecolor{Gray}{gray}{0.9}
\definecolor{newcolor}{rgb}{.8,.349,.1}

\begin{document}
\title{ESA: Annotation-Efficient Active Learning for Semantic Segmentation}
\author{\fontsize{9.5}{4}\selectfont Jinchao Ge$^{1}$, Zeyu Zhang$^{2}$, Minh Hieu Phan$^{1}$, Bowen Zhang$^{1}$, Akide Liu$^{3}$, Yang Zhao$^{4}$, Shuwen Zhao$^{5}$\thanks{Corresponding author: \url{DonutZsw@gmail.com}}}
\authorrunning{Jinchao Ge et al.}
\institute{$^{1}$The University of Adelaide
$^{2}$The Australian National University\\
$^{3}$Monash University
$^{4}$La Trobe University
$^{5}$Tianjin University of Technology}
\maketitle              %
\vspace{-1em}
\begin{abstract}
Active learning enhances annotation efficiency by selecting the most revealing samples for labeling, thereby reducing reliance on extensive human input. Previous methods in semantic segmentation have centered on individual pixels or small areas, 
neglecting the rich patterns in natural images and the power of advanced pre-trained models. To address these challenges, we propose three key contributions: Firstly, we introduce  \textbf{Entity-Superpixel Annotation} (\textbf{ESA}), an innovative and efficient active learning strategy which utilizes a class-agnostic mask proposal network coupled with super-pixel grouping to capture local structural cues. Additionally, our method selects a subset of entities within each image of the target domain, prioritizing superpixels with high entropy to ensure comprehensive representation. Simultaneously, it focuses on a limited number of key entities, thereby optimizing for efficiency. By utilizing an annotator-friendly design that capitalizes on the inherent structure of images, our approach significantly outperforms existing pixel-based methods, achieving superior results with minimal queries, specifically reducing click cost by \textbf{98\%} and enhancing performance by \textbf{1.71\%}. For instance, our technique requires a mere 40 clicks for annotation, a stark contrast to the 5000 clicks demanded by conventional methods.
The code will be available at \url{https://github.com/jinchaogjc/ESA}.

\end{abstract}
\nopagebreak
\vspace{-2em}
\section{Introduction}

\begin{figure}[!htbp]
	\centering
	\begin{subfigure}{0.45\linewidth}
		\centering
		\includegraphics[width=0.99\linewidth]{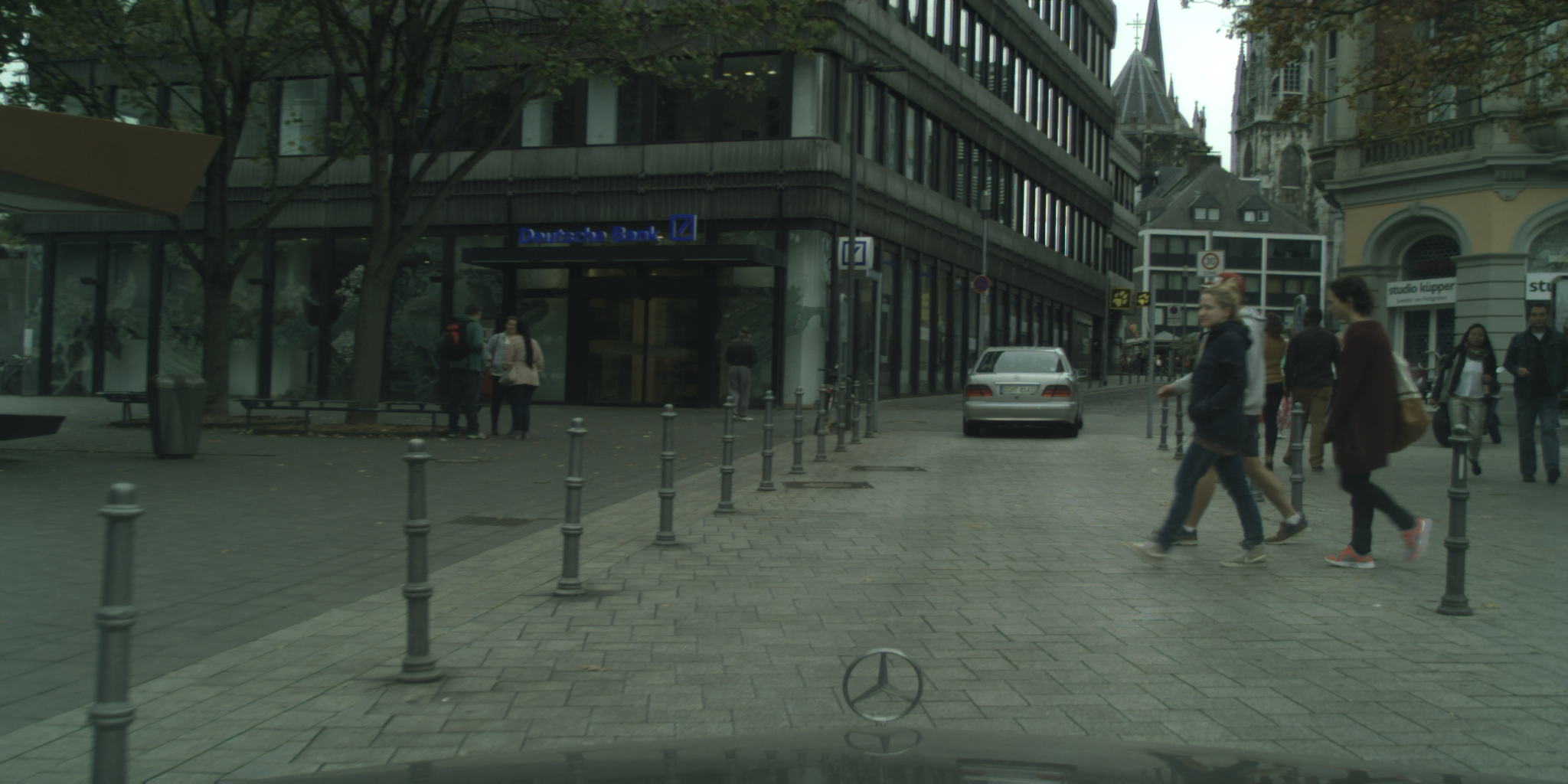}
		\caption{Target image}
		\label{fig:1}%
	\end{subfigure}
	\begin{subfigure}{0.45\linewidth}
		\centering
		\includegraphics[width=0.99\linewidth]{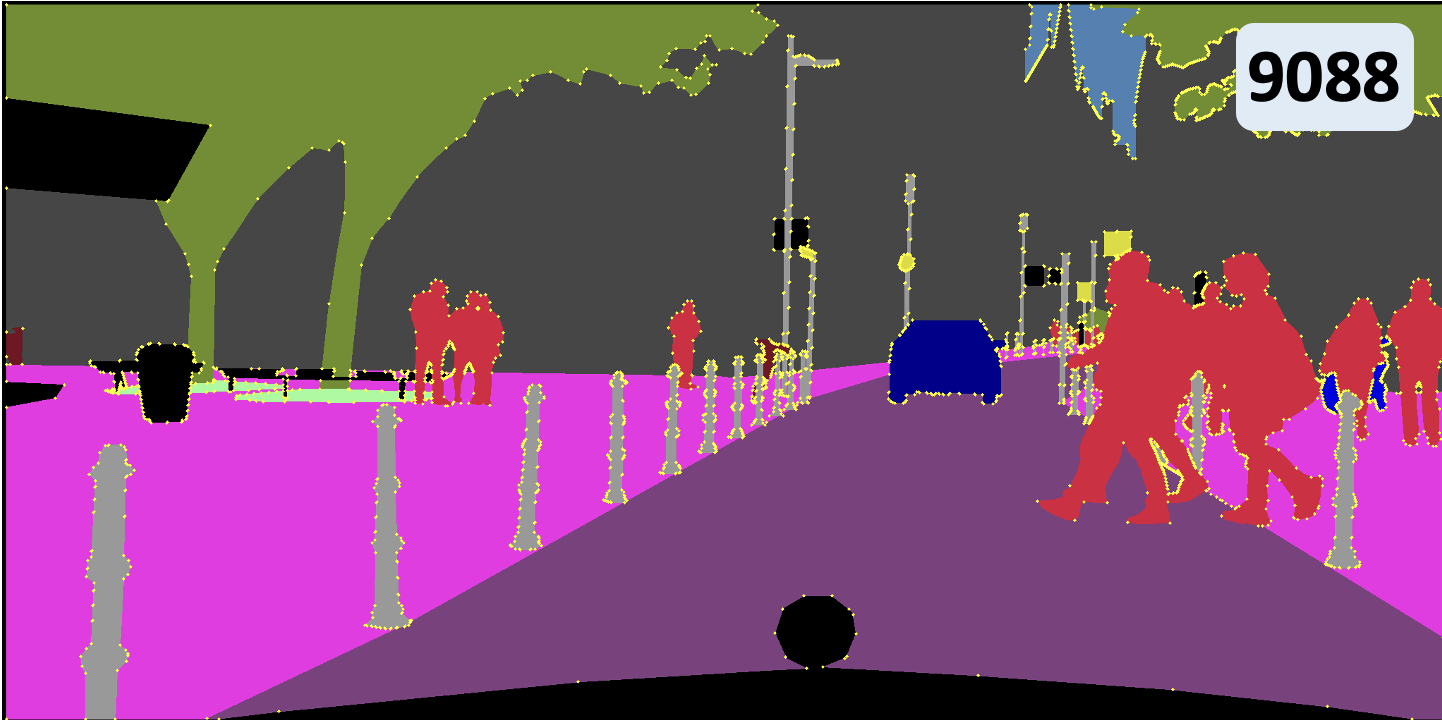}
		\caption{Image-based selection}
		\label{fig:2}
	\end{subfigure}
	\qquad
		\begin{subfigure}{0.45\linewidth}
		\centering
		\includegraphics[width=0.99\linewidth]{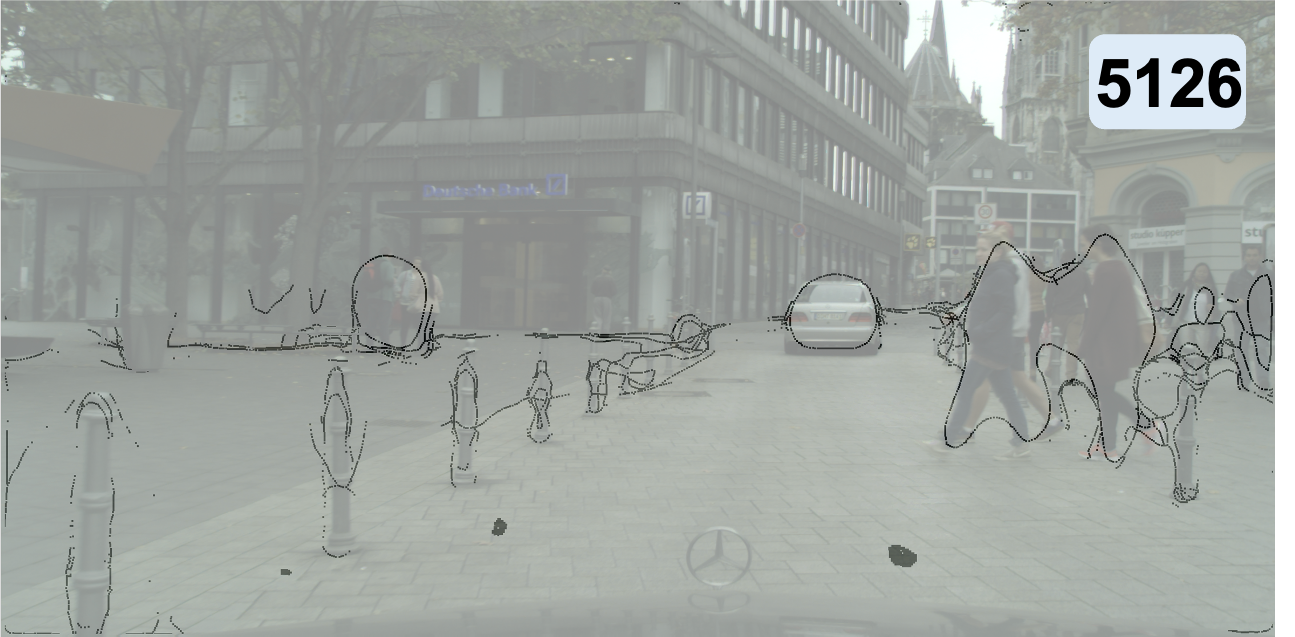}
		\caption{Region-based selection}
		\label{fig:3}
	\end{subfigure}
	\begin{subfigure}{0.45\linewidth}
		\centering
		\includegraphics[width=0.99\linewidth]{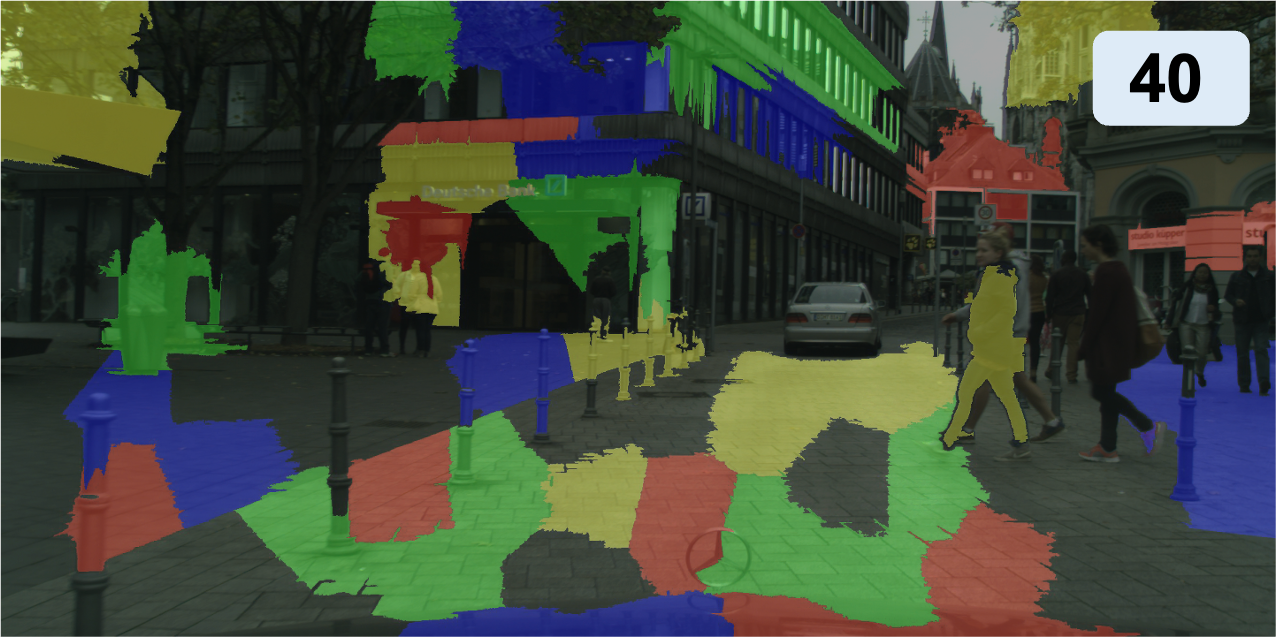}
		\caption{Our selection}
		\label{fig:4}
	\end{subfigure}
	\caption{Comparative analysis of annotation queries by selection strategy. \textbf{(a)} The image is designated for segmentation. \textbf{(b)} The entire image must be annotated~\cite{li2020attention}. \textbf{(c)} Each query encompasses a 3x3 grid, corresponding to a total area of 2.2\% of the entire image~\cite{xie2022towards} \textbf{(d)} Our ESA method selects superpixels and entities for annotation. \textit{The number of queries needed by the annotator for each image is indicated in the upper right corner of figures b), c), and d).}}
    \vspace{-2em}
	\label{fig: annotation cost}
\end{figure}

Semantic segmentation \cite{wei2018revisiting, long2015fully} is essential for a variety of applications such as autonomous driving \cite{lin2016efficient}, medical imaging \cite{wu2023bhsd,zhang2023thinthick,zhang2023segreg,tan2024segstitch}, and remote sensing \cite{li2021multitask}. It assigns object class labels to each pixel in an image, providing a detailed understanding of visual scenes. However, pixel-level annotation of images is a demanding and costly task, requiring substantial human labor. To address this challenge, researchers have turned to active learning, a technique that promises to cut down annotation costs. Active learning selects the most informative samples from a vast array of unlabeled data for annotation, thus minimizing the need for manual effort. Moreover, active learning has been particularly impactful in the domain adaptation of semantic segmentation. It employs pre-trained models from a source domain to identify and query pixels that are uncertain in new target domains. By doing so, it effectively adapts the models to novel visual contexts while keeping the annotation workload to a minimum.

Despite recent progress in active learning for semantic segmentation, two key challenges remain. First, current methods still impose high annotation costs, as illustrated in Fig~\ref{fig: annotation cost}. Take RIPU's region-based approach~\cite{xie2022towards}, for example. It selects a mere 2.2\% or 5\% of an image for annotation but fails to consider the click-based annotation cost. To label a single image under a 2.2\% budget from the Cityscapes dataset~\cite{cordts2016cityscapes}, RIPU~\cite{xie2022towards} requires annotators to make around 5,000 queries, which is nearly as laborious as the traditional labeling methods requiring times to get the labels for one image under a 2.2\% budget of the Cityscapes dataset, which is as time-consuming as labeling the entire image using traditional ``rectangle + polygon"-based labeling method requiring about 9,000 clicks. Second, existing active learning methods mainly target virtual domains populated in synthetic data. For example, some methods~\cite{sharma2021labor} are tested on GTAV~\cite{Richter_2016_ECCV}, a simulated dataset that mimics a gaming environment, and may not accurately reflect real-world conditions. Regrettably, active learning methods in semantic segmentation continue to have several shortcomings. They often overlook the varying difficulty in annotating different semantic areas, such as small or thin objects, which can lead to underperformance for these specific categories.

A significant hurdle in current active learning methods for semantic segmentation is their high annotation costs, which are not time-consuming and expensive. To mitigate these costs, we introduce a novel approach that integrates entity-based with superpixel-based selection to identify the most informative regions for labelling. Our method accesses annotation cost in terms of clicks, strategically choosing regions that offer the greatest information gain per click. This strategy substantially decreases the number of necessary queries and clicks, rendering semantic segmentation more feasible and accessible across a broader spectrum of applications. Additionally, we explore innovative active learning techniques for domain adaptation, utilizing a more general object dataset: COCO common objects~\cite{lin2014microsoft}. Our approach diverges from conventional methods that concentrate on a virtual domain with synthetic data, opting instead of transferring knowledge from a source dataset to a target dataset using authentic real-world data. However, models trained on source domains are susceptible to overfitting, potentially leading to subpar generalization on the target domain \cite{zhang2024jointvit}. Domain adaptation is crucial, as it diminishes the model's dependency on irrelevant domain classes and enhances its generalization capability within the target domain's class subset. Our proposed framework addresses these challenges by selecting samples that are both uncertain and difficult, which we believe will elevate higher performance in complex semantic areas. Moreover, we capitalize unlabeled data for both uncertainty estimation and model training, ensuring a more efficient utilization of the existing data resources.

Our approach to semantic segmentation starts by generating superpixels and entity segmentations across the entire image, followed by the deployment of our active learning algorithm to pinpoint the most informative superpixels. We introduce a novel approach called Entity-Superpixel Annotation (ESA), which integrates the Superpixel-based Annotation (SA) with the selectivity of Entity-based Annotation (EA). The ESA method strategically selects a subset of entities within each image in the target domain for detailed annotation. It labels every superpixel within these selected regions, ensuring a thorough representation, while the EA method complements this by focusing on a limited number of key entities, optimizing for efficiency. To enhance the quality of the learned representations, we've incorporated a novel loss term that is attuned to superpixels. This innovative approach effectively narrows the domain gap between the source and target domains. Additionally, we tackle the challenge of high annotation costs by proposing a class-agnostic mask proposal that is rich in information, directing the annotation efforts towards regions of significant content. As a result, we achieve a more efficient annotation process, minimizing the effort while maximizing the informative value. 

Our work advances the field of active domain adaptation for semantic segmentation through the following contributions:
\vspace{-1em}
\begin{itemize}

\item[\textbullet] We introduce a new active learning method for semantic segmentation called \textbf{Entity-Superpixel Annotation (ESA)}. ESA leverages both superpixels and entities to craft labeling proposals that are pivotal for domain-adaptive semantic segmentation. ESA demands a markedly lower number of queries compared to previous methods. Additionally, we present two foundational processes integral to our methodology: Superpixel-based Annotation (SA) and Entity-based Annotation (EA).

\item[\textbullet] We advocate for the application of domain adaptation (DA) techniques to real-world data, thereby broadening the scope of semantic segmentation. Until now, the focus has predominantly been on street scenes, utilizing synthetic datasets. Our approach, tested on a real-world dataset comprising common objects, establishes new benchmarks for real-data domain adaptations. This marks a significant expansion of the potential application scenarios beyond the traditional scenes. To our knowledge, this is the inaugural study to incorporate realistic scenes in the realm of active learning for semantic segmentation.

\item[\textbullet] Our experimental findings demonstrate that when integrated with DeepLab-v3+, our proposed ESA method achieves a notable enhancement in the performance on domain adaptation benchmarks, exemplified by the COCO $\rightarrow$ VOC transition. Compared to the Pixel-based Annotation (PA) method, ESA achieves a remarkable reduction in click cost by \textbf{98\%}, while also boosting performance by \textbf{1.71\%}.
\end{itemize}

\vspace{-1em}
\section{Related Work}
\vspace{-0.5em}
\noindent \textbf{Superpixels Generation} contains two main approaches: graph-based and clustering-based methods. Superpixels serve as a fundamental, low-level image representation that consolidates visually similar pixels into coherent segments. Graph-based methods conceptualize an image as a graph and employ algorithms such as Normalized Cuts~\cite{ren2003learning}, FH~\cite{felzenszwalb2004efficient}, and ERS~\cite{liu2011entropy}. These algorithms excel at identifying natural boundaries while maintaining a compact and uniform structure within the superpixels. On the other hand, clustering-based methods harness clustering techniques to aggregate pixels. Prominent examples include SLIC~\cite{achanta2012slic}, SEEDS~\cite{van2014seeds}, and LSC~\cite{li2015superpixel}. These methods are particularly adept at capturing fine details and generating smooth boundaries. For the purpose of this study, we have elected to utilize the SLIC~\cite{achanta2012slic} algorithm.

\noindent \textbf{Entity Segmentation}, a concept introduced by
Qi et al.~\cite{qi2022open}, entails the segmentation of all visual entities in an image, differentiating between instance (things) and non-instance (stuffs), without the need to predict their semantic labels. The method extends the capabilities of FCOS~\cite{tian2019fcos}, a widely recognized one-stage detector, enabling the uniform detection of various entities within a scene. The approach is particularly effective in open-world scenarios, where it is not necessary to predefine all possible classes of objects. Instead, it gracefully handles presence of unknown or outlier by incorporating them into the segmentation process. This strategy eliminates the need for outlier spotting in the embedding space, streamlining the detection workflow and enhancing its adaptability to diverse visual environments.

\noindent \textbf{Active Learning (AL)}\cite{peng2024prototype, ning2023madav2} seeks to strike a balance between minimizing annotation costs and maximizing a model's effectiveness. Current strategies in this field can be broadly categorized into three approaches, each with its unique method of selecting data for labeling: diversity-based~\citep{agarwal2020contextual}, uncertainty-based~\cite{shen2017deep}, and hybrid techniques~\cite{yang2017suggestive}. Diversity-based methods works by selecting a range of samples from the unlabeled data that are representative of the whole. They achieve this either by clustering samples based on their feature attribute or by identifying a core set of samples that capture the essence of the data's variability. Uncertainty-based methods, on the other hand, identify the most ambiguous samples that the current model finds most challenging to label. They choose the uncertainty in the model's predictions, using metrics such as entropy, the confidence levels of predictions, and variance in the probabilities predicted by different models within an ensemble. Hybrid sampling leverages the uncertainty and similarity information from FCNs to suggest the most effective areas for annotation, thereby potentially offering a more balanced approach to AL.

\noindent \textbf{Active Learning in semantic segmentation} Semantic segmentation is a critical task in image analysis, and active learning can significantly improve its efficiency. In this domain, annotation methods are typically categorized based on the granularity of the image area being labeled: whole image~\cite{sinha2019vadv}, superpixel~\cite{cai2021superpixel}, polygon~\cite{mittal2019illusions}, and pixel-wise~\cite{shin2021pixelwise} approaches. While various annotation strategies exist for semantic segmentation, their effectiveness is often evaluated without considering the actual annotation cost. This oversight is crucial because the expense of detailed, pixel-level annotation can be prohibitive. Despite its potential advantages, active learning for segmentation has not been extensively explored, largely due to the high cost associated with dense, pixel-level labeling. However, recent research have sought to mitigate this issue by leveraging domain adaptation techniques to transfer models from synthetic dataset to real-world scenarios. For instance, a study by Xie et al.~\cite{xie2022towards} did not fully account for the practical aspects of annotation, such as the number of user clicks required. Their findings suggest that the benefits of using region-based selection methods without additional processing are limited. This raise questions about effectiveness of superpixel-based approach~\cite{cai2021revisiting} in reducing annotation costs, especially when compared to the more traditional "rectangle+polygon" selection methods. Our research is designed to bridge this gap by delving into the potential of superpixel-based approach techniques to minimize the costs associated with semantic segmentation.

\noindent \textbf{Domain adaptation}~\cite{peng2024prototype} is a widely used strategy in machine learning designed to enhance model performance when applied to target domains. The core objective of this technique is to address the challenge of domain shift, a phenomenon where the statistical properties of the source and target domains differ. This discrepancy can lead to models that perform exceptionally well on the source domain but fail to deliver good results on the target domain, a common issue in practical applications where gathering extensive labeled data for the target domain may be costly and impractical. Previous research in active domain adaptation has centered on classification tasks. These studies have attempted to integrate elements of uncertainty and diversity\cite{xie2022towards, cai2021revisiting}, as well as to apply point-based methods~\cite{sharma2021labor}. However, these approaches may sometimes fall short in efficiency or neglect the significance of spatial relationships within data. Our methods leverage superpixels and entities to pinpoint areas of uncertainty for labeling. We aim to elevate model accuracy while simultaneously cutting down on annotation expenses.

\section{Methodology}
\begin{figure*}[t]
    \centering   
    \includegraphics[width=1.0\textwidth]{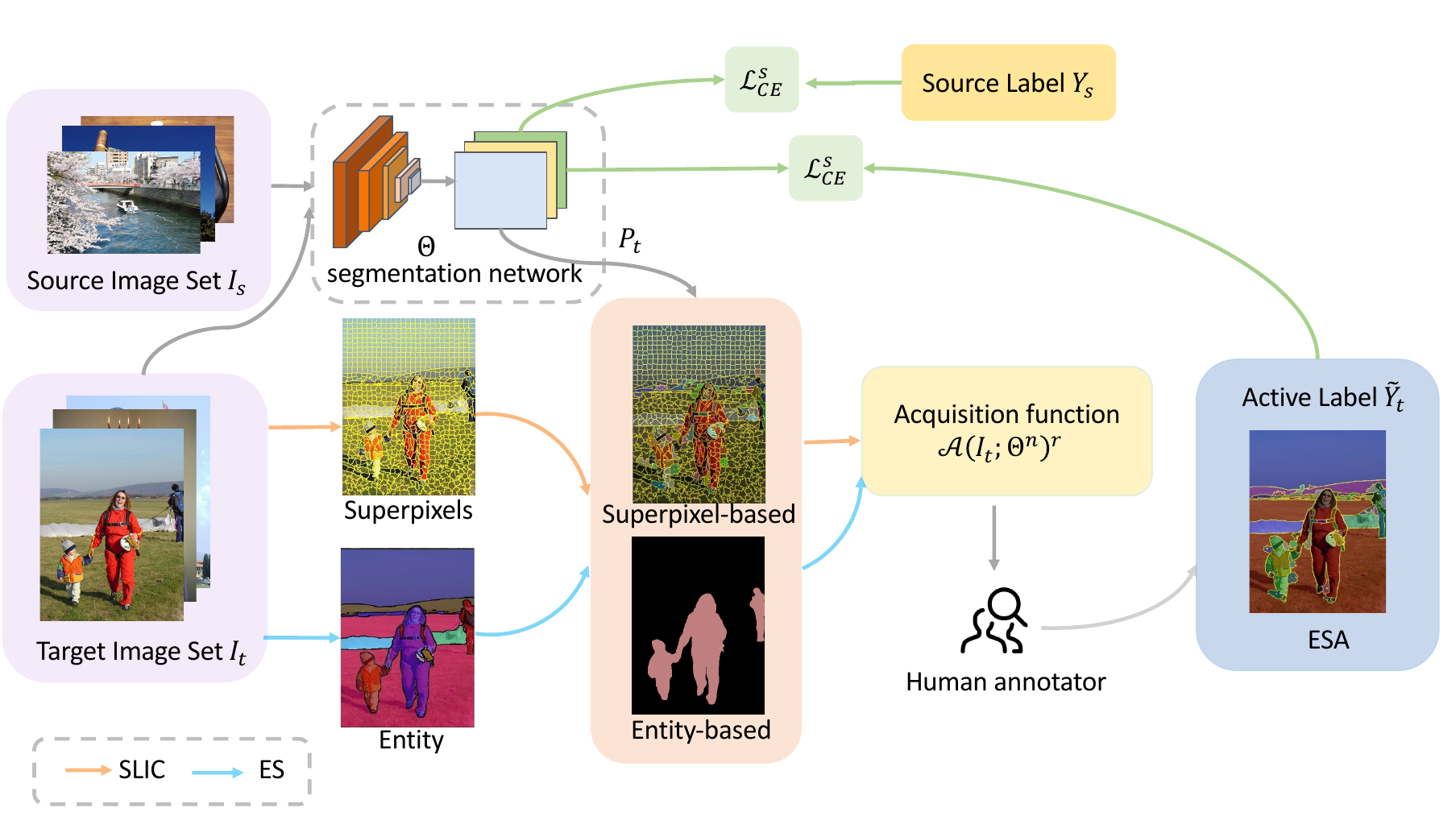} \vspace{-2em}
    \caption{{\bf Active Learning Framework Overview.} We first access the impurity of superpixels and uncertainty of entities within each unlabeled target using the current model. We use an acquisition function $\mathcal{A}$ to select a strategic batch of entities and superpixels for future analysis. With the newly acquired annotation, we then retrain our network, incorporating all previously labeled data to enhance its performance. This cycle continues, progressively refining our model through active selection and human input. } 
    \vspace{-4mm}
    \label{fig:framework}
\end{figure*}

This section provides detailed content of our proposed method for active learning in semantic segmentation tasks, which comprises three key components: superpixel generation (Sec. 3.1), entity generation (Sec. 3.2), and active learning selection (Sec. 3.3). 

\vspace{-1em}
\subsection{Task Formulation}
In our approach to active domain adaptation for semantic segmentation, we utilize two datasets: a \textbf{labeled source dataset} $D_s$ and an \textbf{unlabeled target dataset} $D_t$. The source dataset consists of image-label pairs ${(I_s, Y_s)}$, where $Y_s$ provides pixel-wise annotations for one of $C$ predefined classes within the label set. In contrast, the target dataset includes image-label pairs ${(I_t,\widetilde{Y}_t)}$, with $\widetilde{Y}_t$ initially set to an ignored label, represented by the value 255, indicating that these labels are yet to be determined. Our primary goal is to strategically select informative labels from $\widetilde{Y}_t$ that will enhance the model's performance across target domains.

\subsection{Framework}
ESA's framework is outlined in Fig.~\ref{fig:framework}. It begins with the superpixel generation, which clusters pixels into perceptually coherent superpixel regions that respect object boundaries. This produces results in more natural segments than traditional rectangular approaches. The Entity Selection step further distinguishes distinct visuals entities from the image, providing meaningful components for analysis. ESA considers both the predictive uncertainty within superpixels and the spatial diversity of predictions, guiding the selection of the most informative superpixels for annotation. These selected regions, once annotated, are merged with the source data to fine-tune the semantic segmentation network, optimizing it for superior performance in the target domain. The core of our method is the active learning selection process. The selection is critical for guiding the model training, ensuring that each iteration incorporates the most valuable data. We continue to retrain the model with the annotated samples and the labeled source dataset, refining the model's performance until our annotation budget is fully utilized.

\vspace{-1em}
\subsection{Entity-Superpixel Annotation}
\label{sec: esa method}
Previous region-based active learning approaches in semantic segmentation, such as the RIPU method~\cite{xie2022towards}, commonly segment images into non-overlapping small regions, like 3x3 rectangles. These methods have notable limitations. Specifically, the RIPU method can be inefficient, requiring a multitude of queries to nearly fully annotate an image, which is labor-intensive and time-consuming. In contrast, our Entity-Superpixel Annotation (ESA) considers superpixels and entities as regions, thereby significantly reducing the number of queries needed for annotation. Our approach leverages the inherent structure within images, allowing for a more efficient annotation process. Moreover, ESA can adapt to various image divisions, making it a flexible and general solution for active learning in semantic segmentation.

We begin by introducing two distinct strategies for selecting informative parts of each image, namely ``Superpixel-based Annotation (SA)'' and ``Entity-based Annotation (EA)''. These strategies serve as the foundation for our method, guiding the selection of candidates for annotation.

We initiate our method by generating superpixels using the SLIC algorithm~\cite{achanta2012slic}, an efficient k-means clustering technique that preserves object boundaries to generate $\mathcal{M}_{sa} \in \mathbb{R}^{H \times W \times S}$, where $H$, $W$, and $S$ correspond to the image height, width, and superpixel numbers, respectively. Subsequently, we leverage open world Entity Segmentation (ES)~\cite{qi2022open} to delineate visual entities in the target image $\mathcal{M}_{ea} \in \mathbb{R}^{H \times W \times E}$, where $E$ correspond to the image height, width, and entity numbers, respectively. By adopting the methodology of Qi et al.~\cite{qi2022open}, we generate a set of entity proposals, refining each to produce accurate entity masks $\mathcal{M}_{ea}$.

\noindent \textbf{acquisition function} In our ESA method, a region is defined by a superpixel or entity-shaped mask, which serves as a unit for the annotator to query. Formally, for any region $r$ within an image $\mathbf{I}_t \in \mathbb{R}^{H \times W}$, a selection results $\mathcal{S}$ with the region $\mathcal{M}(i,j)$ is denoted as:

\vspace{-0.5em}
\begin{equation}
    \label{eq:selection}
    \mathcal{S}=\left\{
    \begin{aligned}
        & \mathcal{M}_{sa}(r) \; \text{if SA} \\
        & \mathcal{M}_{ea}(r) \; \text{else EA} \\
    \end{aligned}
    \right.
    ,
    i = \underset{r \notin \widetilde{Y}_t}{\arg \max} \mathcal{A}(\mathbf{I}_t; \Theta^n)^{r}
\end{equation}

Central to our framework is the acquisition function $\mathcal{A}(\mathbf{I}_t; \Theta^n)$, for region r (e.g. superpixel and entities), defined as:
\begin{equation}
    \label{eq:acquisition}
    \mathcal{A}(\mathbf{I}_t; \Theta^n) = \overline{\mathbf{S}}_{t}{(i, j)}
\end{equation}
This function measures uncertainty and confidence, guiding the selection process in the $n^{th}$ iteration, where $\Theta^n$ represents the model.
Our acquisition function is uniquely engineered to capture the unpredictability of superpixel predictions, considering both the spatial variability and the predictive confidence. It intelligently balances these factors to identify samples that significantly contribute to the training dataset, fortifying the model's learning capabilities.

Given a target image $\mathbf{I}_t \in \mathbb{R}^{H\times W}$ and a neural network $\Theta$, the network outputs a softmax prediction $\mathbf{P}_{t} \in \mathbb{R}^{H\times W\times C}$, $C$ corresponding to the channel counts. To quantify prediction uncertainty, we employ the predictive entropy $\mathcal{H}{(i,j)}$ for each pixel, calculated as:
\begin{equation}
    \mathcal{H}{(i, j)} = - \mathbf{P}_{t}{(i, j, c)} \log \mathbf{P}_{t}{(i, j, c)}
\end{equation}
For Superpixels-based Annotation and Entity-based Annotation, the uncertainty of a region is the average of pixel entropy within that region.

Utilizing these predictions, we calculate the mean probability value for each superpixel, denoted as $\overline{\mathbf{S}}_{t}{(i, j)}$, using the following equation:
\begin{equation}
    \overline{\mathbf{S}}_{t}{(i, j)}  =\frac{1}{\left|\mathcal{M}{(i, j)}\right|} \sum_{r \in \mathcal{M}{(i, j)}} \mathcal{H}{(i, j)}
\end{equation}
This equation averages the predicted probability values of all pixels within a superpixel to obtain its mean probability value.

\vspace{-1em}
\subsubsection*{Selection Process}
Our refinement of the selection process prioritizes enhancing both the quality and the informative content of our training dataset. We establish the selection criteria based on the acquisition scores, giving precedence to superpixels that achieve the highest scores, while ensuring non-redundancy of these selections in relation to previously chosen entities. The target pseudo label is derived by selecting the output with the maximum probability:
\begin{equation}
    \label{eq:selection_process}
    \widetilde{\mathbf{Y}}_{t}{(i, j)} = \mathop{\arg\max}_{l \in {1,...,L}} \overline{\mathbf{S}_{t}}{(i, j)}
\end{equation}

Our ESA method boosts the efficiency of the annotation process by strategically identifying regions that require  review, all while minimizing disruption. This targeted approach is crucial for lightening the workload on annotators. As a result, it enables a more practical and cost-effective active learning strategy for semantic segmentation.

\vspace{-0.5em}
\subsection{Training Object}
To address the task of refining our network through a balanced approach that considers both target and source domains, we employ a tailored loss function. This function is crafted to integrate a supervised loss component with a negative loss term:
\begin{equation}
    \min_{\Theta} \mathcal{L}_{sup} +  \alpha\mathcal{L}_{n}^t\,.
        \label{eq:over_all_loss}
\end{equation}
By default, the weighting coefficients $\alpha$ are set to 1.0 respectively in all experiments.

Firstly, we use the labeled data from the source domain to provide a robust foundation for our network's learning process. This initial phase establishes a comprehensive understanding of the general features present across domains. Secondly, we augment this foundation with a focused annotation effort on superpixels and entities within the target domain. This targeted enhancement allows the network to absorb and adapt to the difference that are specific to the target domain, thus enhancing its overall discriminatory capabilities. In order to do this, all labeled data from both the source and target domains are used to fine-tune the network. This is achieved by optimizing the standard supervised loss function:
\begin{equation}
        \mathcal{L}_{sup} = \mathcal{L}_{CE}^s(\mathbf{I}_s, \mathbf{Y}_s) + \mathcal{L}_{CE}^t(\mathbf{I}_t,\widetilde{\mathbf{Y}}_t) \,,
\end{equation}
where $\mathcal{L}_{CE}$ is the categorical cross-entropy (CE) loss:

\begin{equation}
    \label{eq:loss_seg}
    \mathcal{L}_{CE} = -\frac{1}{|\mathbf{I}|}\sum_{(i,j) \in \mathbf{I}} \sum_{c=1}^C  \mathbf{Y}{(i, j, c)}\log \mathbf{P}{(i, j, c)}\,
\end{equation}

where $\mathbf{Y}{(i, j, c)}$ denotes the label for pixel $(i,j)$.

Observe that the negative pseudo labels for the negative instances are binary. Thus, we define the loss function for negative learning as follows: 
\begin{equation}
    \mathcal{L}_{n}^t  = \frac{-1}{\sum_{(i,j)\in \mathbf{I}_t} \sum_{c=1}^C \Pi } \sum_{(i,j)\in \mathbf{I}_t} \sum_{c=1}^C \Pi \log (1-\mathbf{P}_{t}{(i, j, c)}) \label{eq:loss_neg} \,
\end{equation}

Here 
$\Pi=\pi(\mathbf{P}_{t}{(i, j, c)})$. $\pi(\mathbf{P}_{t}{(i, j, c)})=1$ when 
$\mathbf{P}_{t}{(i, j, c)} > \tau$, otherwise, $\pi(\mathbf{P}_{t}{(i, j, c)})=0$. We set $\tau=0.05$

\subsection{Algorithm}
Algorithm~\ref{alg:ESA} outlines our iterative approach to train models on source data while selectively annotating target data, optimizing performance with acquisition functions and budget constraints

\begin{algorithm}[t]
  \DontPrintSemicolon
  {\bf Require}: Source dataset with annotation $(\mathbf{I}_s, \mathbf{Y}_s)$, target dataset without label $\mathbf{I}_t$, total iteration $N$, selection rounds $S$, per-round budget $b$, and hyperparameters: $\tau$, $\alpha$.\\
  {\bf Define:} define empty target selection sets $\widetilde{\mathbf{Y}}_{t}=\emptyset$\\
  Get model parameters $\Theta^{0}$ trained on source dataset with Eq.~\ref{eq:loss_seg}.\\
  \For{$n=1 \text{ to } N$}
  {
      Train the model $\Theta^{n}$ via Eq.~\ref{eq:over_all_loss} with a batch of source samples and target samples. \\
      \If{$n \in S$}
      {
            Compute acquisition scores in $\mathbf{I}_t$ based on Eq.~\ref{eq:acquisition}. \\
            Sample superpixels or entities using Eq~\ref{eq:selection} and annotate them by Eq~\ref{eq:selection_process}
            until the per-round budget $b$ is exhausted. Add new annotated samples to set $\widetilde{\mathbf{Y}}_{t}$.\\
        }
  }
  {\bf Return:} Model parameters $\Theta^{N}$.
  \caption{Our proposed methods} 
  \label{alg:ESA}
\end{algorithm} 

\vspace{-1em}
\section{Experiments}
\vspace{-0.5em}

\noindent \textbf{Dataset}. 
In this study, we conduct experiments to transfer real-world data of common objects from COCO dataset, which contains 11k images with 81 classes, to VOC dataset, which contains 1464 images with 21 classes.

\noindent \textbf{Implementation details.} 
We adapt DeepLab-v3+ and ResNet-101 as model backbones on Tesla V100 GPU using PyTorch. Our models are optimized using the SGD optimizer with a momentum of 0.9, weight decay of 0.0005, and poly learning rate schedule with the initial learning rate set to 0.00025. The training process is stopped after 40k iterations with batch size 2. Both the source and target datasets are resized to a uniform dimension of, 512$\times$512 to maintain consistency during training. For Superpixel-based Annotation (SA), we set the parameters k to 1000 and compactness to 0.1. Furthermore, we have reimplemented RIPU~\cite{xie2022towards} as our baseline.

\noindent \textbf{Evaluation metric}. The evaluation metric used is mean Intersection-over-Union (mIoU) on the VOC validation set.

\noindent \textbf{Annotation budget} The selection process for annotation budget involves five iterative rounds, with a total of 40 superpixels per image selected for SA and entity scores larger than 50\% selected for Entity-based Annotation (EA). The detailed annotation budget is shown in Tab.~\ref{table:coco2voc_active}. It is worth noting that previous pixel-wise and local region-based methods require far more clicks for annotation to densely label each image. In contrast, our ESA approach leverages perceptually grouped regions as the units for annotation, significantly reducing the manual effort and cost required to provide comprehensive image labels during the active learning process.

\subsection{Main Results}

\vspace{-2em}
\begin{table*}[ht]
  \centering
  \caption{\textbf{Comparison with previous results on task COCO $\rightarrow$ VOC.} Detailed results are presented in Sec~\ref{sec:main results}. `*'~denotes~re-implemented~baselines~for~COCO~$\rightarrow$~VOC~task.
    Abbreviated class names: 
    \textit{back.}~=~background, 
    \textit{aero.}~=~aeroplane, 
    \textit{bott.}~=~bottle, 
    \textit{moto.}~=~motorbike, 
    \textit{pers.}~=~person, 
    \textit{plant}~=~potted~plant, 
    \textit{tv}~=~TV/monitor. 
    Full class names: bicycle, bird, boat, bus, car, cat, chair, cow, dining table, dog, horse, sheep, sofa, train.}
  \label{table:coco2voc_active}
  \resizebox{0.9\textwidth}{!}{
  \begin{threeparttable}
  \begin{tabular}{l c c c c c c c c c c c c }
  \toprule[1.2pt]
  \midrule
  \multirow{2}{*}{Method} & \rotatebox{60}{back.} & \rotatebox{60}{aero.} & \rotatebox{60}{bike} & \rotatebox{60}{bird} & \rotatebox{60}{boat} & \rotatebox{60}{bott.} & \rotatebox{60}{bus} & \rotatebox{60}{car} & \rotatebox{60}{cat} & \rotatebox{60}{chair} & \rotatebox{60}{cow} \\
  \cmidrule{2-12}
   & \rotatebox{60}{table} & \rotatebox{60}{dog} & \rotatebox{60}{horse} & \rotatebox{60}{moto.} & \rotatebox{60}{pers.} & \rotatebox{60}{plant} & \rotatebox{60}{sheep} & \rotatebox{60}{sofa} & \rotatebox{60}{train} & \rotatebox{60}{tv} & \rotatebox{60}{mIoU} \\
  \midrule
 Source Only & 91.16 & 86.18 & 37.25 & 80.09 & 57.99 & 45.64 & 87.43 & 78.63 & 86.29 & 28.94 & 70.66 \\ 
            & 37.46 & 84.28 & 64.61 & 71.83 & 83.36 & 38.25 & 76.19 & 40.19 & 72.00 & 71.91 &  66.21 \\
  \midrule
  WDA$^{\star}$(Image)~\citep{paul2020domain} & 91.26 & 82.7 & 30.73 & 80.13 & 66.12 & 49.93 & 92.77 & 69.17 & 84.99 & 33.21 & 88.82 \\
            & 52.66 & 80.42 & 81.89 & 75.79 & 81.31 & 41.60 & 79.53 & 50.29 & 82.84 & 63.87 & 69.53 \\
  WDA$^{\star}$(Point)~\citep{paul2020domain} & 91.76 & 81.03 & 32.31 & 76.71 & 67.24 & 53.29 & 93.07 & 72.23 & 87.38 & 33.11  & 87.71 \\
            & 58.32 & 79.95 & 79.3 & 73.98 & 83.3 & 44.31 & 81.13 & 53.61 & 81.38 & 61.6 & 70.13 \\
  RIPU$^{\star}$ (PA 40)~\citep{xie2022towards} & 93.60 & 89.74 & 51.65 & 83.48 & 71.81 & 65.96 & 91.32 & 83.77 & 90.46 & 28.09 & 72.77 \\
            & 57.70 & 82.14 & 75.38 & 82.16 & 84.94 & 35.86 & 73.56 & 44.99 & 78.97 & 72.07 & 71.93 \\
  \bf \cellcolor{Gray} Ours (SA 40) &  \bf \cellcolor{Gray} 92.61 &  \bf \cellcolor{Gray} 86.82 &  \bf \cellcolor{Gray} 42.85 &  \bf \cellcolor{Gray} 83.37 &  \bf \cellcolor{Gray} 73.87 &  \bf \cellcolor{Gray} 59.96 &  \bf \cellcolor{Gray} 89.89 &  \bf \cellcolor{Gray} 84.07 &  \bf \cellcolor{Gray} 87.85 &  \bf \cellcolor{Gray} 36.15 &  \bf \cellcolor{Gray} 83.11 \\
    \bf \cellcolor{Gray}  &  \bf \cellcolor{Gray} 59.71 &  \bf \cellcolor{Gray} 81.12 &  \bf \cellcolor{Gray} 77.86 &  \bf \cellcolor{Gray} 80.62 &  \bf \cellcolor{Gray} 83.93 &  \bf \cellcolor{Gray} 54.40 &  \bf \cellcolor{Gray} 82.02 &  \bf \cellcolor{Gray} 47.66 &  \bf \cellcolor{Gray} 82.19 &  \bf \cellcolor{Gray} 74.31 &  \bf \cellcolor{Gray} 73.54\\
  RIPU$^{\star}$ (PA 1000)~\citep{xie2022towards} & 93.67 & 87.94 & 55.08 & 83.53 & 68.32 & 61.67 & 90.37 & 81.86 & 87.64 & 29.85 & 74.09 \\
            & 55.76 & 82.87 & 75.15 & 78.64 & 85.35 & 52.60 & 68.02 & 51.01 & 82.15 & 75.17 & 72.42 \\
  
  \bf \cellcolor{Gray} Ours (SA 1000) &  \bf \cellcolor{Gray} 93.87 &  \bf \cellcolor{Gray} 88.73 &  \bf \cellcolor{Gray} 54.14 &  \bf \cellcolor{Gray} 87.01 &  \bf \cellcolor{Gray} 74.84 &  \bf \cellcolor{Gray} 72.20 &  \bf \cellcolor{Gray} 88.05 &  \bf \cellcolor{Gray} 85.43 &  \bf \cellcolor{Gray} 90.49 &  \bf \cellcolor{Gray} 35.33 &  \bf \cellcolor{Gray} 73.07 \\
   \bf \cellcolor{Gray}          &  \bf \cellcolor{Gray} 54.53 &  \bf \cellcolor{Gray} 86.53 &  \bf \cellcolor{Gray} 72.94 &  \bf \cellcolor{Gray} 81.10 &  \bf \cellcolor{Gray} 85.85 &  \bf \cellcolor{Gray} 50.42 &  \bf \cellcolor{Gray} 82.99 &  \bf \cellcolor{Gray} 51.74 &  \bf \cellcolor{Gray} 80.19 &  \bf \cellcolor{Gray} 75.52 &  \bf \cellcolor{Gray} 74.52\\

  \midrule[1.2pt]

   \bf \cellcolor{Gray} Ours (EA 33) &  \bf \cellcolor{Gray} 93.11 &  \bf \cellcolor{Gray} 83.59 &  \bf \cellcolor{Gray} 42.44 &  \bf \cellcolor{Gray} 88.84 &  \bf \cellcolor{Gray} 67.34 &  \bf \cellcolor{Gray} 67.90 &  \bf \cellcolor{Gray} 91.16 &  \bf \cellcolor{Gray} 82.33 &  \bf \cellcolor{Gray} 89.94 &  \bf \cellcolor{Gray} 28.93 &  \bf \cellcolor{Gray} 77.98 \\
    \bf \cellcolor{Gray}         &  \bf \cellcolor{Gray} 56.45 &  \bf \cellcolor{Gray} 84.50 &  \bf \cellcolor{Gray} 77.15 &  \bf \cellcolor{Gray} 79.97 &  \bf \cellcolor{Gray} 85.10 &  \bf \cellcolor{Gray} 52.95 &  \bf \cellcolor{Gray} 79.12 &  \bf \cellcolor{Gray} 47.68 &  \bf \cellcolor{Gray} 78.85 &  \bf \cellcolor{Gray} 75.77&  \bf \cellcolor{Gray} 72.91\\
  \midrule[1.2pt]
  RIPU$^{\star}$ (PA, 5767) ~\citep{xie2022towards} & 93.13 & 83.18 & 46.9 & 84.62 & 69.47 & 61.77 & 87.68 & 84.8 & 88.52 & 36.19 & 80.17 \\
            & 57.86 & 84.22 & 80.54 & 73.08 & 84.82 & 45.70 & 82.36 & 46.17 & 73.01 & 76.53 & 72.41\\
  RIPU$^{\star}$ (RA, 640) ~\citep{xie2022towards} & 93.96 & 91.80 & 56.53 & 87.25 & 74.78 & 68.43 & 84.80 & 85.64 & 88.69 & 37.25 & 82.05 \\
            & 48.85 & 84.90 & 78.14 & 81.96 & 86.03 & 53.19 & 81.90 & 49.04 & 78.47 & 74.76 & 74.69 \\
  
  \midrule[1.2pt]
  
  \bf \cellcolor{Gray} Ours (ESA, 102) &  \bf \cellcolor{Gray} 93.23 &  \bf \cellcolor{Gray} 88.49 &  \bf \cellcolor{Gray} 42.10 &  \bf \cellcolor{Gray} 89.22 &  \bf \cellcolor{Gray} 75.10 &  \bf \cellcolor{Gray} 70.01 &  \bf \cellcolor{Gray} 88.57 &  \bf \cellcolor{Gray} 84.49 &  \bf \cellcolor{Gray} 88.75 &  \bf \cellcolor{Gray} 36.99 &  \bf \cellcolor{Gray} 72.57 \\
   \bf \cellcolor{Gray}          &  \bf \cellcolor{Gray} 56.25 &  \bf \cellcolor{Gray} 84.24 &  \bf \cellcolor{Gray} 77.92 &  \bf \cellcolor{Gray} 82.65 &  \bf \cellcolor{Gray} 84.16 &  \bf \cellcolor{Gray} 53.91 &  \bf \cellcolor{Gray} 79.67 &  \bf \cellcolor{Gray} 49.05 &  \bf \cellcolor{Gray} 83.89 &  \bf \cellcolor{Gray} 75.27 &  \bf \cellcolor{Gray} 74.12\\
  \midrule
  
  Fully Supervised $^{\sharp}$ & 93.63 & 89.49 & 59.32 & 82.60 & 72.89 & 61.93 & 88.23 & 84.77 & 89.18 & 30.72 & 81.13 \\
            & 57.26 & 85.10 & 76.90 & 81.84 & 85.31 & 51.07 & 75.48 & 46.41 & 78.01 & 71.72 & 73.48 \\
  \midrule
  \bottomrule[1.2pt]
  \end{tabular}
  \end{threeparttable}
  }
\end{table*}
\vspace{-1em}

\label{sec:main results}
Our study presents the outcomes of the COCO $\to$ VOC tasks, detailed in Table~\ref{table:coco2voc_active}. 
The ``Source Only'' approach, which involves training exclusively on the COCO dataset and then evaluating the performance on the VOC dataset. Our SA approach demonstrates a significant advantage over the Pixel-based Annotating (PA) method of RIPU~\cite{xie2022towards}. Specifically, with a limited budget of only 40 query time, the SA method shows a 1.61\% improvement in performance. This margin expands to a 2.1\% enhancement when the budget is increased to 1000 query time. Additionally, the segmentation results obtained by our method are characterized by smoother contours, indicating its superior performance.

The EA method demonstrates superior efficiency and performance compared to the PA method. On average, EA requires only 33 queries per image to achieve a mIoU score that is 0.49\% higher than PA method, which necessitates a considerably larger number of 1000 queries per image.

In subsequent experiments, we explored a high-budget annotation scheme, selecting 2.2\% of pixels per image for labeling. The PA methods required 5767 queries, a number that significantly exceeds the requirements of traditional annotation techniques. In comparison, the Region-based Annotating (RA) which defines regions using a $3\times 3$ rectangular area, required a more modest 640 queries. Our proposed method, ESA, not only outperforms the PA method at 2.2\% pixel selection rate with an average of just 102 queries per image but also delivers performance on par with RA. This highlights the effectiveness of ESA in achieving high accuracy with fewer queries, thereby, optimizing both the annotation process and the learning outcome.

Our findings clearly demonstrate that our proposed method outperforms state-of-the-art methods, even with a limited number of budgets per target image. This suggests that active learning could be a promising approach to address domain adaptation challenges.

\vspace{-1em}
\subsection{Qualitative results}

\begin{figure*}[!htbp]
      \centering  
      \includegraphics[width=0.85\textwidth]{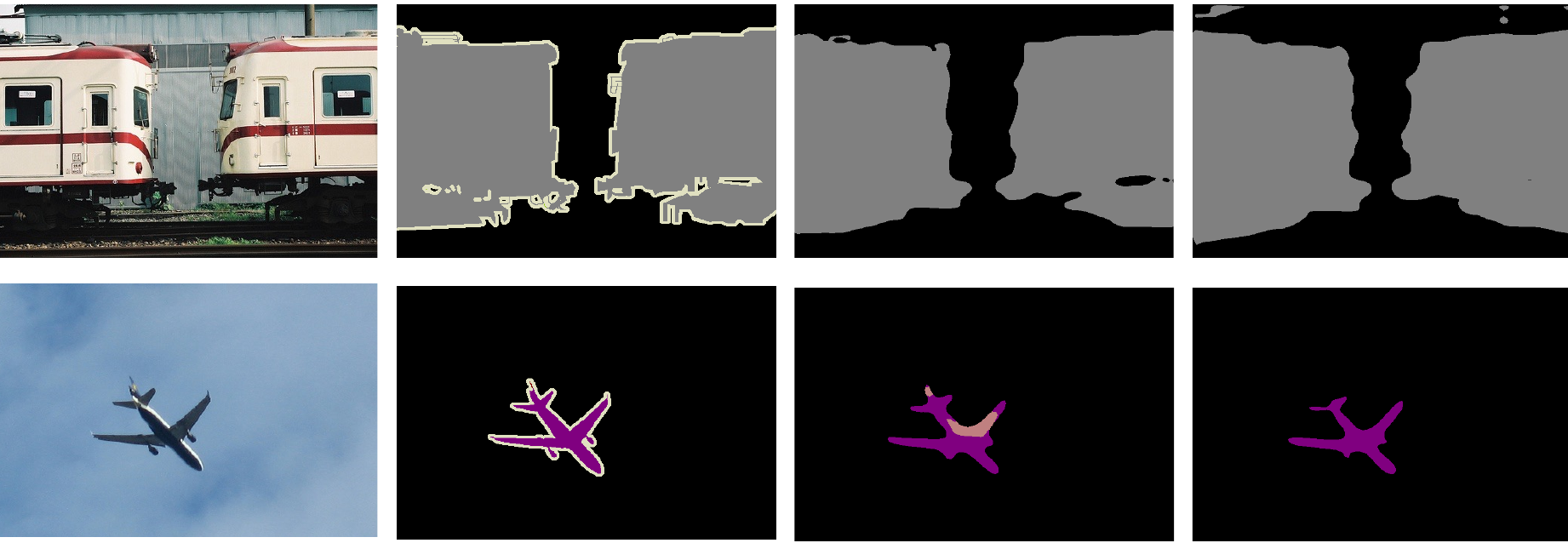}
      \vspace{-2mm}
      \caption{\textbf{Visualization of segmentation results for the task COCO $\to$ VOC.} From left to right: original target image, ground-truth label, PA results predicted by RIPU~\cite{xie2022towards}, results predicted by Ours SA method are shown one by one.}
      \label{fig:visulization_results}
      \vspace{-0.5em}
\end{figure*}

\vspace{-1em}
In Figure~\ref{fig:visulization_results}, we display a visual comparison to evaluate the effectiveness of our proposed method, designated as SA, against the PA method. Our method delivers results of enhanced refinement and fewer artifacts, presenting a smoother visual appearance. This comparison is particularly pertinent as both methods operate within the same query budget of 40 times.

\vspace{-0.5em}
\subsection{Ablation Study}
To evaluate each component of our proposed method, we conducted ablation studies transitioning from the COCO $\rightarrow$ VOC, as shown in Table~\ref{table:ablation},~\ref{table:active_selection_methods}. We select SA method for our ablation study because it shares the same budge constraints as previous studies, allowing for a fair comparison of methodologies.

\begin{table}[t]
  \resizebox{0.99\linewidth}{!}{
  \begin{minipage}{0.5\linewidth}
      \centering
      \caption{\textbf{Ablation study.} (a): use region impurity only as the selection criterion. (b): use prediction uncertainty only as the selection criterion. (c): combine impurity and uncertainty.
      }
      \label{table:ablation}
      \begin{tabular}{c c c c }
      \toprule
       & \multicolumn{2}{c}{Selection} \\
      \midrule
      Method & Impurity & Uncertainty & mIoU\\
      \midrule
      \midrule
      (a) & $\checkmark$ & &  \textbf{73.13}\\
      (b) & & $\checkmark$ &   72.03\\
      (c) & $\checkmark$ & $\checkmark$ & 72.18\\
      \midrule
      \bottomrule
      \end{tabular} 
  \end{minipage}
  \hfill
  \hspace{2em}
  \begin{minipage}{0.5\linewidth}
      \centering
      \caption{Experiments on different \textbf{active selection methods}. Comparison of active selection methods: random selection (RAND), entropy-based selection (ENT), softmax confidence based selection (SCONF), and our method as described in Sec. \ref{sec: esa method}.}
      \label{table:active_selection_methods}
      \begin{tabular}{l c c }
      \toprule
      Method & Budget & mIoU\\
      \midrule
      RAND & 40 pixels & 70.72  \\
      ENT~\citep{shen2017deep} & 40 pixels & 72.18 \\
      SCONF~\citep{culotta2005reducing} & 40 pixels & 72.03  \\
      \bf \cellcolor{Gray}Ours  & \cellcolor{Gray}40 pixels & \bf \cellcolor{Gray}73.54 \\    
      \bottomrule
      \end{tabular} 
  \end{minipage}
  }
\end{table}

\paragraph{Effect of region impurity and prediction uncertainty.} 
To further investigate the efficacy of each component of our, we perform ablation studies on COCO $\to$ VOC. In Table~\ref{table:ablation}, configurations (b) and (c) achieve a clear improvement compared to RAND, indicating that incorporating insights from region impurity and prediction uncertainty effectively identifies valuable image areas. Notably, region impurity shows a more substantial advantage with a 1.1\% improvement compared to configuration (b), likely due to its ability to leverage the spatial adjacency within images, thus mitigating class imbalance issues. Moreover, configuration (a) outperforms both (b) and (c), suggesting that the impurity selection criterion is particularly effective in capturing a diverse range of regions that showing potential for generalization.

\vspace{-1em}
\paragraph{Comparison of different active selection methods.} As illustrated in Tab.~\ref{table:active_selection_methods}, our PA method surpasses alternative approaches, yielding consistent and noteworthy improvements. To delve deeper into the performance benefits conferred by our selection strategy, we compare SA with established selection methods including random selection (RAND), entropy-based selection (ENT)~\cite{shen2017deep} and softmax confidence based selection (SCONF)~\cite{culotta2005reducing}, in scenarios devoid of the $\mathcal{L}_{n}^t$ term. Constrained by a stringent pixel budget of 40, both ENT and SCONF suffer from pixel aggregation, which results in performance degradation. In contrast, ours SA secures a substantial enhancement of 2.82\% mIoU over RAND, underscoring the efficacy of our method in navigating the limitations imposed by pixel redundancy and optimizing the use of constrained pixel budget. The consistent increments observed across the 40 query budget further substantiate the superiority and distinct advantages of our approach.

\paragraph{Effect of $\mathcal{L}_{n}^t$.} Our ablation studies reveal the impact of our proposed $\mathcal{L}_{n}^t$ on performance. When $\mathcal{L}_{n}^t$ is removed, we observe a comparable decline in performance. Ultimately, our method, which incorporated $\mathcal{L}_{n}^t$, yields better results, highlighting the advantages of our tailored loss functions.

The incremental and consistent improvements from the SA method to our comprehensive approach substantiate and efficacy of each element. SA encapsulates the basis of our approach, making it a representative prototype for analysis. The insights gleaned from SA are foundational and have guided the refinement of our other methods, which share its core principles. Then, a focused examination of SA allows for an in-depth understanding of the incremental improvements each component brings to the system. In the supplement section, we will show more visualizations that illustrate its predictive impact and a comparative analysis with SoTA domain adaptation losses.

\begin{spacing}{1.0}
    \section{Conclusion}
Our study aims to evaluate the efficiency of superpixel-based or entity-based selection strategies for active domain adaptation in semantic segmentation of common object datasets, using a performance benchmarking analysis. Our findings expose the limitations of these strategies, suggesting that adapting real-world common object data by refining a broader class spectrum into more focused categories is a more realistic approach than relying on synthesis data. We propose a novel superpixel-entity based acquisition strategy, termed ''ESA'', which utilizes both superpixels and entities to generate labeling proposals. Experimental results address that the ``ESA'' significantly outperformed existing methods across adaptation benchmarks, exemplified by the COCO $\rightarrow$ VOC adaptation task. These results indicate the potential of superpixel-entity based annotation strategies for improving domain adaptive semantic segmentation within complex datasets.

\end{spacing}

\end{document}